\documentclass[11pt,a4paper]{article}
\usepackage{authblk}
\usepackage[hyperref]{acl2019}
\usepackage{times}
\usepackage{latexsym}

\usepackage{url}

\usepackage{graphicx}
\usepackage[english]{babel}
\usepackage{blindtext}
\usepackage{amsfonts}
\usepackage{amsmath}
\usepackage{url}
\usepackage{array}
\usepackage{comment}
\usepackage{amssymb}
\usepackage{todonotes}
\usepackage[hang,flushmargin]{footmisc}
\usepackage{footmisc}
\usepackage{enumitem}

\aclfinalcopy 
\def\aclpaperid{1851} 

\newcommand\smallleq{\scalebox{0.9}{$\leq$}}

\title{Robust Zero-Shot Cross-Domain Slot Filling with Example Values}

\author[1]{\textbf{Darsh J Shah*}}
\author[2]{\textbf{Raghav Gupta*}}
\author[2]{\textbf{Amir A Fayazi}}
\author[3]{\textbf{Dilek Hakkani-T{\"u}r}}
\affil[1]{MIT CSAIL, Cambridge, MA}
\affil[2]{Google Research, Mountain View, CA}
\affil[3]{Amazon Alexa AI, Sunnyvale, CA}
\affil[ ]{\texttt{darsh@csail.mit.edu, \{raghavgupta,amiraf\}@google.com,}}
\affil[ ]{\texttt{dilek@ieee.org}}

\begin{document}
\setlength{\abovedisplayskip}{5pt}
\setlength{\belowdisplayskip}{5pt}
\setlength{\abovedisplayshortskip}{5pt}
\setlength{\belowdisplayshortskip}{5pt}
\maketitle
\begin{abstract}
Task-oriented dialog systems increasingly rely on deep learning-based slot filling models, usually needing extensive labeled training data for target domains. Often, however, little to no target domain training data may be available, or the training and target domain schemas may be misaligned, as is common for web forms on similar websites. Prior zero-shot slot filling models use slot descriptions to learn concepts, but are not robust to misaligned schemas. We propose utilizing both the slot description and a small number of examples of slot values, which may be easily available, to learn semantic representations of slots which are transferable across domains and robust to misaligned schemas. Our approach outperforms state-of-the-art models on two multi-domain datasets, especially in the low-data setting.
\end{abstract}
\let\svthefootnote\thefootnote
\let\thefootnote\relax\footnote{Asterisk (\textbf{*}) denotes equal contribution. Research conducted when all authors were at Google Research.}
\addtocounter{footnote}{-1}
\let\thefootnote\svthefootnote
\vspace{-3.5mm}
\section{Introduction}
Goal-oriented dialog systems assist users with tasks such as finding flights, booking restaurants and, more recently, navigating user interfaces, through natural language interactions. Slot filling models, which identify task-specific parameters/slots (e.g. \textit{flight date}, \textit{cuisine}) from user utterances, are key to the underlying spoken language understanding (SLU) systems. Advances in SLU have enabled virtual assistants such as Siri, Alexa and Google Assistant. There is also significant interest in adding third-party functionality to these assistants. However, supervised slot fillers \cite{young2002talking, bellegarda2014spoken} require abundant labeled training data, more so with deep learning enhancing accuracy at the cost of being data intensive \cite{mesnil2015using,kurata2016leveraging}.

\begin{figure}
    \centering
    \includegraphics[width=0.47\textwidth]{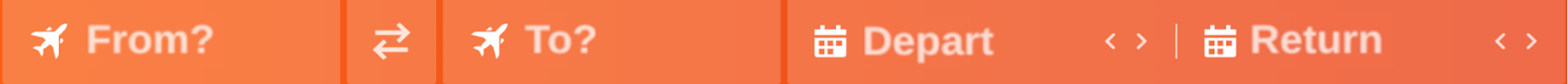}
      \\[8pt]
    \includegraphics[width=0.47\textwidth]{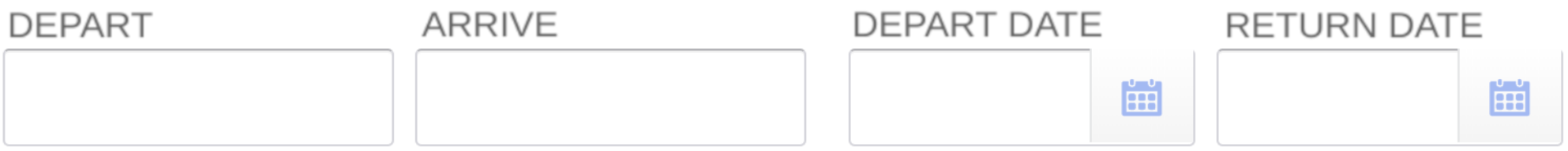}
    \caption{Misaligned schemas for flight booking from \url{kayak.com} (top) and \url{southwest.com} (bottom): 
     slot name \textit{depart} in the two schemas refers to departure date and departure city respectively, hence models trained on one schema may falter on the other.
     }
    \label{fig:flight_screenshot}
    \vspace{-15pt}
\end{figure}

Two key challenges with scaling slot fillers to new domains are adaptation and misaligned schemas (here, slot name mismatches). 
Extent of supervision may vary across domains: there may be ample data for \textit{Flights} but none for \textit{Hotels}, requiring models to leverage the former to learn semantics of reusable slots (e.g. \textit{time}, \textit{destination}).
In addition, schemas for overlapping domains may be incompatible by way of using different names for the same slot or the same name for different slots. This is common with web form filling: two sites in the same domain may have misaligned schemas, as in Figure \ref{fig:flight_screenshot}, precluding approaches that rely on schema alignment. 

\begin{figure*}[t!]
    \centering
    \includegraphics[width=\textwidth]{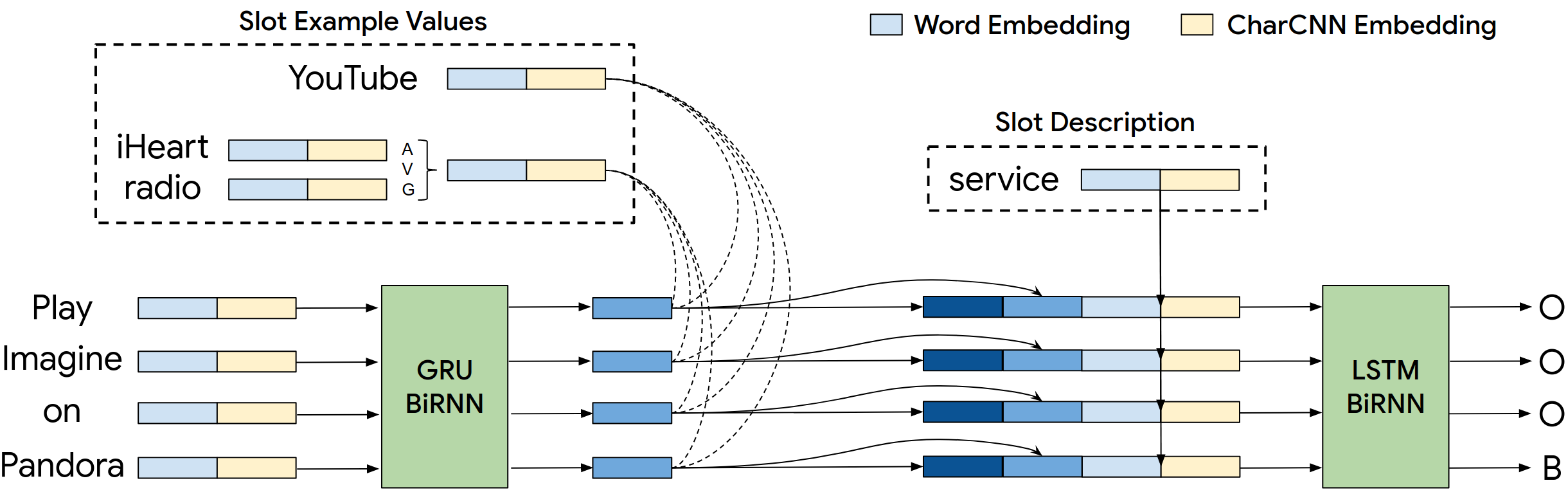}
    \caption{Illustration of the overall model with all inputs and outputs shown.}
    \label{fig:model_fig}
    \vspace{-3pt}
\end{figure*}

Zero-shot slot filling, typically, either relies on slot names to bootstrap to new slots, which may be insufficient for cases like in Figure \ref{fig:flight_screenshot}, or uses hard-to-build domain ontologies/gazetteers.
We counter that by supplying a small number of example values in addition to the slot description to condition the slot filler. This avoids negative transfer from misaligned schemas and further helps identify unseen slots while retaining cross-domain transfer ability. Besides, example values for slots can either be crawled easily from existing web forms or specified along with the slots, with little overhead.

Given as few as 2 example values per slot, our model surpasses prior work in the zero/few-shot setting on the SNIPS dataset by an absolute 2.9\% slot F1, and is robust to misaligned schemas, as experiments on another multi-domain dataset show.

\section{Related Work}
Settings with resource-poor domains are typically addressed by adapting from resource-rich domains \citep{blitzer2006domain,pan2010cross,guo-etal-2018-multi}. General domain adaptation techniques such as multi-task learning \cite{jaech2016domain, goyal2018fast, siddhant2018unsupervised} and domain adversarial learning  \cite{liu2017multi} have been adapted and applied to SLU and related tasks \cite{henderson2014third}. Work targeting domain adaptation specifically for this area includes, modeling slots as hierarchical concepts \cite{zhu2018concept} and using ensembles of models trained on data-rich domains \cite{gavsic2015policy, kim2017domain, jha2018bag}.\\

Zero-shot learning \citep{norouzi2013zero,socher2013zero} by way of task descriptions has gained popularity, and is of interest for data-poor settings like ours. Particularly, work on zero-shot utterance intent detection has leveraged varied resources like click logs \cite{dauphin2013zero} and manually-defined domain ontologies \cite{kumar2017zero}, as well as models such as deep structured semantic models \cite{chen2016zero} and capsule networks \cite{xia2018zero}. Zero-shot semantic parsing is addressed in \newcite{krishnamurthy2017neural} and \newcite{herzig2018decoupling} and specifically for SLU utilizing external resources such as label ontologies in \newcite{ferreira2015online, ferreira2015zero} and handwritten intent attributes in \newcite{yazdani2015model}. Our work is closest in spirit to \newcite{bapna2017towards} and \newcite{lee2018zero}, who employ textual slot descriptions to scale to unseen intents/slots. Since slots tend to take semantically similar values across utterances, we augment our model with example values, which are easier to define than manual alignments \cite{li2011multi}.

\section{Problem Statement}

\begin{figure}[h!]
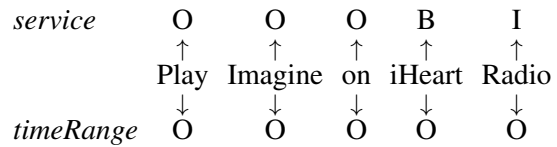

\setlength\tabcolsep{3.5pt}
\def\arraystretch{0.75}
\setlength{\abovecaptionskip}{5pt}
\setlength{\belowcaptionskip}{-5pt}
\newcommand{\smalluparrow}{\small$\uparrow$\normalsize}
\newcommand{\smalldownarrow}{\small$\downarrow$\normalsize}
\centering
  \begin{tabular}{l c c c c c}\textit{service}&  O & O & O & B & I\\
& \smalluparrow & \smalluparrow & \smalluparrow & \smalluparrow & \smalluparrow \\
 & Play & Imagine & on & iHeart & Radio\\
 & \smalldownarrow & \smalldownarrow & \smalldownarrow & \smalldownarrow & \smalldownarrow \\
\textit{timeRange} & O & O & O & O & O\\
\end{tabular}
  \caption{Example semantic frame with
  $IOB$ slot annotations for a positive and a negative instance.}
  \label{fig:clu_example}
\end{figure}

We frame our conditional sequence tagging task as follows: given a user utterance with $T$ tokens and a slot type, we predict inside-outside-begin (IOB) tags $\{y_1, y_2\ldots y_T\}$ using 3-way classification per token, based on if and where the provided slot type occurs in the utterance. Figure \ref{fig:clu_example} shows IOB tag sequences for one positive (slot \textit{service}, present in the utterance) and one negative (slot \textit{timeRange}, not present in the utterance) instance each.

\begin{table*}[ht]
\small
\def\arraystretch{1.16}
\setlength\tabcolsep{4pt}
\centering
\resizebox{2.05\columnwidth}{!}{
    \begin{tabular}[t]{|l|>{\raggedright}p{12.85cm}|}
    \hline \textbf{Intent} & \textbf{Slot Names (Training and Evaluation)}\tabularnewline 
    \hline
      \small AddToPlaylist & artist,  entityName, musicItem, playlist, playlistOwner \tabularnewline
 \small BookRestaurant &  city, cuisine, partySizeNumber, restaurantName, restaurantType, servedDish, spatialRelation, state\ldots  \tabularnewline
 \small GetWeather & \small city, conditionDescription, country, geographicPoi, spatialRelation, state, timeRange\ldots \tabularnewline
 \small PlayMusic & \small album, artist, genre, musicItem, playlist, service, sort, track, year  \tabularnewline
 \small RateBook & \small bestRating, objectName, objectPartOfSeriesType, objectSelect, objectType, ratingUnit, ratingValue  \tabularnewline
 \small SearchCreativeWork & \small objectName, objectType  \tabularnewline
 \small FindScreeningEvent & \small locationName, movieName, movieType, objectLocationType, objectType, spatialRelation, timeRange
 \tabularnewline
 \hline
    \end{tabular}
    }
    \\[9.5pt]
\resizebox{2.05\columnwidth}{!}{
    \begin{tabular}[t]{|l|>{\raggedright}p{5.89cm}|>{\raggedright}p{7.96cm}|}
    \hline \textbf{Intent} & \textbf{Training Slot Names} & \textbf{Evaluation Slot Names}\tabularnewline \hline
BookBus & from, to, leaving, returning, travelers, tripType, departureTime & from, to, departOn, addReturnTrip, tripType, promoCode, discountOptions, children, adults, seniors \tabularnewline
FindFlights & from, to, depart, return, cabinClass, flightType & depart, arrive, departDate, returnDate, searchType, promoCode \tabularnewline
BookRoom & where, checkIn, checkOut, guests, homeType, propertyType, priceRange, amenities & location, hotelName, checkIn, checkOut, rooms, roomType, pricePerNight, rating, amenities \tabularnewline 
\hline
    \end{tabular}
    }
    \caption{Intents and training/evaluation slot schemas for SNIPS (top) and XSchema (bottom) datasets.}
     \label{table:datasets}
     \vspace{-10pt}
\end{table*}

\section{Model Architecture}
Figure \ref{fig:model_fig} illustrates our model architecture where a user utterance is tagged for a provided slot. To represent the input slot, along with a textual slot description as in \newcite{bapna2017towards}, we supply a small set of example values for this slot, to provide a more complete semantic representation.\footnote{Note that the slot description is still needed since example slot values alone cannot distinguish slots which take semantically similar values (e.g. \textit{departDate} vs \textit{returnDate}).} Detailed descriptions of each component follow. \\[-9pt] 

\noindent\textbf{Inputs:} We use as input $d_{wc}$-dimensional embeddings for 3 input types: $T$ user utterance tokens $\{u_i \in \mathbb{R}^{d_{wc}},\, 1 \smallleq i \smallleq T\}$, $S$ input slot description tokens  $\{d_i \in \mathbb{R}^{d_{wc}},\, 1 \smallleq i \smallleq S\}$, and $K$ example values for the slot, with the $N_k$ token embedding for the $k^{th}$ example denoted by $\{e^k_i \in \mathbb{R}^{d_{wc}},\, 1 \smallleq i \smallleq N_k\}$.\\[-9pt] 

\noindent\textbf{Utterance encoder:} We encode the user utterance using a $d_{en}$-dimensional bidirectional GRU recurrent neural network (RNN) \cite{chung2014empirical}. We denote the set of per-token RNN hidden states by $H = \{h_i \in \mathbb{R}^{d_{en}},\, 1 \smallleq i \smallleq T\}$, which are used as contextual utterance token encodings.

\vspace{-12pt}
\begin{gather}
H = \scalebox{0.85}{$BiGRU$}(\{u_i,\, 1 \smallleq i \smallleq T\})
\end{gather}

\noindent\textbf{Slot description encoder:} We obtain an encoding $d^{s} \in \mathbb{R}^{d_{wc}}$ of the slot description by mean-pooling the embeddings for the $S$ slot description tokens.
\begin{gather}
d^{s} = \frac{1}{S}\sum_{i=1}^Sd_i 
\end{gather}

\noindent\textbf{Slot example encoder:} We first obtain encodings $\{e_k^{x} \in \mathbb{R}^{d_{wc}},\, 1 \smallleq k \smallleq K\}$ for each slot example value by mean-pooling the $N_k$ token embeddings. Then, we compute an attention weighted encoding of all $K$ slot examples $\{e^{a}_i \in \mathbb{R}^{d_{wc}},\, i\smallleq 1 \smallleq T\}$ for each utterance token, with the utterance token encoding as attention context. Here, $\alpha^{x}_{i} \in \mathbb{R}^K$ denotes attention weights over all $K$ slot examples corresponding to the $i^{th}$ utterance token, obtained with general cosine similarity \cite{luong2015effective}.
\begin{gather}
e_k^{x} = \frac{1}{N_k}\sum_{i=1}^{N_k}e^k_i,\, 1 \smallleq k \smallleq K\\
\alpha^{x}_{i} = \scalebox{0.85}{$softmax$}(\{h_i \mathbf{W}_{a} e_k^{x}\, \forall k\}), 1\smallleq i \smallleq T\\
e^{a}_i = \sum_{k=1}^K \alpha^{x}_{i_k} \times e_k^{x}
\end{gather}

\noindent\textbf{Tagger:} We feed the concatenated utterance, slot description and example encodings to a $d_{en}$-dimensional bidirectional LSTM. 
The output hidden states $X = \{x_i \in \mathbb{R}^{d_{en}}, 1\smallleq i \smallleq T\}$ are used for a 3-way IOB tag classification per token.
\begin{gather}
X = \scalebox{0.85}{$BiLSTM$}(\{h_i \oplus d^{s} \oplus e^{a}_i,\, 1\smallleq i \smallleq T\})\\
y_i = \scalebox{0.85}{$softmax$}(\mathbf{W}_tx_i + \mathbf{b}_t), 1\smallleq i \smallleq T
\end{gather}

\noindent\textbf{Parameters:} We use fixed $d_{w}{=}128$-dim pretrained word embeddings\footnote{\small\texttt{https://tfhub.dev/google/nnlm-en-dim128/1}} for all tokens. We also train per-character embeddings, fed to a 2-layer convolutional neural network \cite{kim2014convolutional}  to get a $d_{c}{=}32$-dim token embedding. For all inputs, the $d_{wc}{=}160$-dim final embedding is the concatenation of the word and char-CNN embeddings. The RNN encoders have hidden state size $d_{en}{=}128$. All trainable weights are shared across intents and slots. The model relies largely on fixed word embeddings to generalize to new intents/slots.

\section{Datasets and Experiments}\vspace{-2.5pt}
In this section we describe the datasets used for evaluation, baselines compared against, and more details on the experimental setup. \\[-16pt]

\paragraph{Datasets:}
In order to evaluate cross-domain transfer learning ability and robustness to misaligned schemas, respectively, we use the following two SLU datasets to evaluate all models. 
\begin{itemize}[leftmargin=*]
    \item \textbf{SNIPS}: This is a public SLU dataset
    \cite{coucke2018snips}
    of crowdsourced user utterances with 39 slots across 7 intents and ${\sim}2000$ training instances per intent. Since 11 of these slots are shared (see Table \ref{table:datasets}), we use this dataset to evaluate cross-domain transfer learning.
    \vspace{-2mm}
    \item \textbf{XSchema}: This is an in-house crowdsourced dataset with 3 intents ($500$ training instances each). Training and evaluation utterances are annotated with different schemas (Table \ref{table:datasets}) from real web forms to simulate misaligned schemas. 
\end{itemize}

\paragraph{Baselines:} We compare with two strong zero-shot baselines: Zero-shot Adaptive Transfer (ZAT) \cite{lee2018zero} and Concept Tagger (CT) \cite{bapna2017towards}, in addition to a 2-layer multi-domain bidirectional LSTM baseline \cite{dilek:2016} for non-zero-shot setups. ZAT and CT condition slot filling only on slot descriptions, with ZAT adding slot description attention, char embeddings and CRFs on top of CT. Since labor-intensive long text descriptions are unavailable for our data, we use tokenized slot names in their place, as in \newcite{bapna2017towards}.\\[-9pt]

\paragraph{Experimental Setup:} 

\label{sec:experimental_setup}
We use SNIPS to test zero/few-shot 
transfer: for each target intent $I$, we train on all ${\sim}2000$ training instances from intents other than $I$, and varying amounts of training data for $I$, evaluating exclusively on $I$. For XSchema, we train and evaluate on a single intent, specifically evaluating cross-schema performance.

\begin{table}[t!]
\footnotesize
\setlength\tabcolsep{1.8pt}
\def\arraystretch{1.23}
\centering
\resizebox{\columnwidth}{!}{
    \begin{tabular}[t]{|l | c c c | c c c c |}\hline
    Target training e.g. & \multicolumn{3}{c|}{0} &\multicolumn{4}{c|}{50}\\\hline
    Intent $\downarrow$ Model $\rightarrow$ & CT & ZAT & +2Ex  & LSTM & CT & ZAT & +10Ex  \\\hline
    AddToPlaylist & 53.3 &	46.8 &	\textbf{55.2} &  59.4 &	74.4  & 73.4 & \textbf{76.2*} \\
    BookRestaurant & 45.7 &	46.6 & \textbf{48.6*} &  57.5 &	\textbf{63.8}	& 63.5 & 63.6  \\
    GetWeather & 63.5 &	60.7 &	\textbf{66.0*}	&  75.7 & 72.1 &	71.1 & \textbf{77.5*}  \\
    PlayMusic & 28.7 &	30.1 &	\textbf{33.8*}  &  49.3	& 56.4 &	56.0 & \textbf{58.8} \\
    RateBook & 24.5 &	\textbf{31.0} & 28.5 &  \textbf{85.1*} & 82.9 &	83.8 & 82.2  \\
    SearchCreativeWork & 24.7	 & \textbf{26.7} &	26.2 &  52.9 &	62.8 &	63.7 &  \textbf{65.9} \\
    FindScreeningEvent & 23.7	& 19.7 & \textbf{25.5*} &  60.8 & 64.9 & 64.6 &  \textbf{67.0*} \\\hline
    Average & 37.7	& 37.4 &	\textbf{40.6*} &  62.8 &	68.2  &	68.0 & \textbf{70.1*} \\\hline
    \end{tabular}
    }
    \caption{Slot F1 scores for baselines (CT, ZAT, LSTM) and our best models (with 2 slot values for zero-shot and 10 values for 50 train instances) on SNIPS. Rows represent different train-test splits, defined in Section \ref{sec:experimental_setup}. Our model consistently outperforms the baselines, with ${\sim}3\%$ absolute gain in the zero-shot setting.\protect\footnotemark}
    \label{table:table_results}
\end{table}

We sample positive and negative instances (Figure \ref{fig:clu_example}) in a ratio of 1:3. Slot values input during training and evaluation are randomly picked from values taken by the input slot in the relevant domain's training set, excluding ones that are also present in the evaluation set. In practice, it is usually easy to obtain such example values for each slot either using automated methods (such as crawling from existing web forms) or have them be provided as part of the slot definition, with negligible extra effort.

To improve performance on out-of-vocabulary entity names, we randomly replace slot value tokens in utterances and slot examples with a special token, and raise the replacement rate linearly from 0 to 0.3 during training \cite{rastogi2018multi}. 

The final cross-entropy loss, averaged over all utterance tokens, is optimized using ADAM \cite{kingma2014adam} for 150K training steps. Slot F1 score \cite{tjong2000introduction} is our final metric, reported after 3-fold cross-validation. 

\section{Results}
\label{sec:results}
For the SNIPS dataset, Table \ref{table:table_results} shows slot F1 scores for our model trained with randomly-picked slot value examples in addition to slot descriptions vis-\`a-vis the baselines. Our best model consistently betters the zero-shot baselines CT and ZAT, which use only slot descriptions, overall and individually for 5 of 7 intents. The average gain over CT and ZAT is ${\sim}3\%$ in the zero-shot case. In the low-data setting, all zero-shot models gain ${\geq}5\%$ over the multi-domain LSTM baseline (with the 10-example-added model further gaining ${\sim}2\%$ on CT/ZAT). All models are comparable when all target data is used for training, with F1 scores of $87.8\%$ for the LSTM, and $86.9\%$ and $87.2\%$ for CT and our model with 10 examples respectively.

\begin{table}[t!]
\footnotesize
\setlength\tabcolsep{2.0pt}
\def\arraystretch{1.23}
\centering
\resizebox{0.93\columnwidth}{!}{
    \begin{tabular}[t]{|l | c c c | c c c |}\hline
    Target training e.g. & \multicolumn{3}{c|}{0} &\multicolumn{3}{c|}{50}\\\hline
    Intent $\downarrow$ Model $\rightarrow$ & CT & ZAT & +10Ex & CT & ZAT & +10Ex  \\\hline
    BookBus & 70.9 &	70.1 &	\textbf{74.1*} & 86.8 & 85.2 & \textbf{89.4}\\
    FindFlights & 43.5 & 44.8 & \textbf{53.2*} & 62.3 & 59.7 & \textbf{69.2*}\\
    BookRoom & 23.6 & 23.4 & \textbf{33.0*} & 49.7 & 52.1 & \textbf{58.7*}\\\hline
    \end{tabular}
    }
    \caption{Slot F1 scores on the XSchema dataset\protect\footnotemark[4]. We train and evaluate on a single intent, but with different schemas, thus precluding the LSTM baseline.}
    \label{tab:cross_schema_results}
    \vspace{-3pt}
\end{table}

Table \ref{tab:cross_schema_results} shows slot F1 scores for XSchema data. Our model trained with 10 example values is robust to varying schemas, with gains of ${\sim}3\%$ on \textit{BookBus}, and ${\sim}10\%$ on \textit{FindFlights} and \textit{BookRoom} in the zero-shot setting. 

For both datasets, as more training data for the target domain is added, the baselines and our approach perform more similarly. For instance, our approach improves upon the baseline by ${\sim}0.2\%$ on SNIPS when 2000 training examples are used for the target domain, affirming that adding example values does not hurt regular performance.\\[-9pt]

\footnotetext{Asterisk (\textbf{*}) indicates a statistically significant gain over the second-best model as per McNemar's test ($p< 0.05$).}

\noindent \textbf{Results by slot type:} 
Example values help the most with limited-vocabulary slots not encountered during training: our approach gains ${\geq}20\%$ on slots such as \textit{conditionDescription}, \textit{bestRating}, \textit{service} (present in intents \textit{GetWeather}, \textit{RateBook}, \textit{PlayMusic} respectively). Intents \textit{PlayMusic} and \textit{GetWeather}, with several limited-vocabulary slots, see significant gains in the zero-shot setting.

For compositional open-vocabulary slots (\textit{city}, \textit{cuisine}), our model also compares favorably - e.g. $53\%$ vs $27\%$ slot F1 for unseen slot \textit{cuisine} (intent \textit{BookRestaurant}) - since the semantic similarity between entity and possible values is easier to capture than between entity and description.

Slots with open, non-compositional vocabularies (such as \textit{objectName}, \textit{entityName}) are hard to infer from slot descriptions or examples, even if these are seen during training but in other contexts, since utterance patterns are lost across intents. All models are within $5\%$ slot F1 of each other for such slots. This is also observed for unseen open-vocabulary slots in the XSchema dataset (such as \textit{promoCode} and \textit{hotelName}).

\begin{figure}[t!]
\vspace{-4pt}
    \hspace{4pt}
    \includegraphics[width=0.416\textwidth, height=0.32\textwidth]{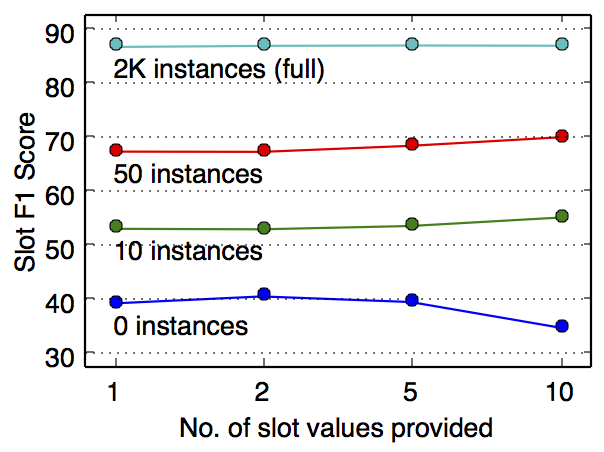}
    \vspace{-2pt}
    \caption{Variation of overall slot F1 score with number of slot value examples input to the model, with varying number of target intent training instances for SNIPS.}
    \label{fig:values_fig}
    \vspace{-5pt}
\end{figure}

For XSchema experiments, our model does significantly better on slots which are confusing across schemas (evidenced by gains of ${>}20\%$ on \textit{depart} in \textit{FindFlights}, \textit{roomType} in \textit{BookRoom}).  \\[-9pt]

\noindent \textbf{Effect of number of examples:} Figure \ref{fig:values_fig} shows the number of slot value examples used versus performance on SNIPS. For the zero-shot case, using 2 example values per slot works best, possibly due to the model attending to perfect matches during training, impeding generalization when more example values are used. In the few-shot and normal-data settings, using more example values helps accuracy, but the gain drops with more target training data. For XSchema, in contrast, adding more example values consistently improves performance, possibly due to more slot name mistmatches in the dataset. We avoid using over 10 example values, in contrast to prior work \cite{krishnamurthy2017neural, naik2018contextual} since it may be infeasible to easily provide or extract a large number of values for unseen slots.\\[-9pt]

\noindent \textbf{Ablation:} Slot replacement offsets overfitting in our model, yielding gains of $2{-}5\%$ for all models incl. baselines. Fine-tuning the pretrained word embeddings and removing character embeddings yielded losses of ${\sim}1\%$. We tried more complex phrase embeddings for the slot description and example values, but since both occur as short phrases in our data, a bag-of-words approach worked well.\\[-9pt]

\noindent \textbf{Comparison with string matching:} A training and evaluation setup including example values for slots may lend itself well to adding string matching-based slot fillers for suitable slots (for example, slots taking numeric values or having a small set of possible values). However, this is not applicable to our exact setting since we ensure that the slot values to be tagged during evaluation are never provided as input during training or evaluation. In addition, it is difficult to distinguish two slots with the same expected semantic type using such an approach, such as for slots \textit{ratingValue} and \textit{bestRating} from SNIPS intent \textit{RateBook}.

\section{Conclusions and Future Work}
We show that extending zero-shot slot filling models to use a small number of easily obtained example values for slots, in addition to textual slot descriptions, is a scalable solution for zero/few-shot slot filling tasks on similar and heterogenous domains, while resistant to misaligned overlapping schemas. Our approach surpasses prior state-of-the-art models on two multi-domain datasets.

The approach can, however, be inefficient for intents with many slots, as well as potentially sacrificing accuracy in case of overlapping predictions. Jointly modeling multiple slots for the task is an interesting future direction. Another possible direction is to incorporate zero-shot entity recognition \cite{guerini2018toward}, thereby eliminating the need for example values during inference.

In addition, since high-quality datasets for downstream tasks in dialogue systems (such as dialogue state tracking and dialogue management) are even more scarce, exploring zero-shot learning approaches to these problems is of immense value in building generalizable dialogue systems.

\section*{Acknowledgements}
We would like to thank Ankur Bapna for the insightful discussions that have notably shaped this work. We would also like to thank the Deep Dialogue team at Google Research for their support.

\bibliography{acl2019}
\bibliographystyle{acl_natbib}
\end{document}